\documentclass{article}

\PassOptionsToPackage{numbers, compress}{natbib}
\usepackage[preprint]{neurips_2026}

\usepackage[utf8]{inputenc} 
\usepackage[T1]{fontenc}    
\usepackage{hyperref}       
\usepackage{url}            
\usepackage{booktabs}       
\usepackage{amsfonts}       
\usepackage{nicefrac}       
\usepackage{microtype}      
\usepackage{xcolor}         
\usepackage{graphicx}
\usepackage{tabularx}
\usepackage{caption}
\usepackage{multicol}
\usepackage{amsmath}
\usepackage{multirow}

\title{Modeling Atomic Conformational Ensembles of Proteins via Test-Time Supervision of Boltz-2 on Cryo-EM Density Maps}

\author{Jay Shenoy \\
Department of Computer Science\\
Stanford University, Flatiron Institute, \& SLAC National Lab \\
Stanford, CA 94305, USA \\
\texttt{jshenoy}@stanford.edu \\
\And
Miro Astore \\
Center for Computational Biology \& Center for Computational Mathematics \\
Flatiron Institute \\
New York, NY 10010, USA \\
\And
Axel Levy\thanks{Work done while at Stanford.} \\
Department of Electrical Engineering \\
Stanford University \& SLAC National Lab \\
Stanford, CA 94305, USA \\
\And
Frédéric Poitevin \\
LCLS Data Systems \\
SLAC National Lab \\
Menlo Park, CA 94025, USA \\
\And
Sonya M. Hanson \\
Center for Computational Biology \& Center for Computational Mathematics \\
Flatiron Institute \\
New York, NY 10010, USA \\
\And
Gordon Wetzstein \\
Department of Electrical Engineering \\
Stanford University \\
Stanford, CA 94305, USA \\
}

\begin{document}

\maketitle

\begin{abstract}
  Knowledge of a protein's atomic conformational ensemble is critical to determining its function, yet state-of-the-art ensemble prediction models are limited by lack of high-quality conformational data from simulation or experiment. Recent advances in heterogeneous reconstruction for cryo-electron microscopy (cryo-EM) have enabled scientists to visualize ensembles of density maps for larger proteins and complexes not typically accessible through simulation, but building atomic models into these maps remains a challenge. Traditionally, ensemble prediction models are trained via a two-stage process: experimental density maps are converted into atomic structural ensembles through model building, after which these structures are used to train sequence-to-atomic ensemble predictors. In this work, we propose a new principle for fine-tuning pre-trained static structure prediction models such as Boltz-2 directly on raw cryo-EM maps, bypassing the two-stage process. We apply this technique to the problem of atomic model building by fine-tuning Boltz-2 to generate atomic conformations from an input ensemble of cryo-EM maps, achieving superior model building accuracy compared to prior work. Beyond overfitting to individual map ensembles, our method, CryoSampler, also shows preliminary evidence of in-domain generalization after fine-tuning, sampling diverse atomic conformations for an unseen sequences within the same protein family without requiring cryo-EM data. These capabilities indicate that CryoSampler holds the potential to train next-generation atomic ensemble prediction models directly on raw cryo-EM measurements.
\end{abstract}

\section{Introduction}
\label{sec:introduction}

Prediction of protein atomic ensembles from sequence is widely regarded as the next frontier in computational structural biology \cite{lane2023protein, sala2023modeling, ourmazd2022structural}, following the landmark success of AlphaFold \cite{jumper2021highly, abramson2024accurate} in the modeling of static structures. Despite this promise, the development of dynamic models has been hindered by the scarcity of high-quality protein ensemble data in public repositories. Today, X-ray crystallography accounts for the majority of the deposited structures in the Protein Data Bank (PDB), but cryo-electron microscopy (cryo-EM) has grown rapidly in popularity in recent years \cite{berman2000protein}. Critically, the imaging conditions of cryo-EM allow proteins to explore a broader range of conformations (shapes) than crystallography permits, making it a natural source of ensemble data. Cryo-EM is therefore a compelling choice for training next-generation ensemble predictors, particularly for large molecular systems where simulation-based approaches face practical size and timescale constraints.

However, integrating cryo-EM data into model training is challenging. Although algorithms can reconstruct ensembles of cryo-EM maps \cite{punjani20213d, zhong2021cryodrgn, cryodrgnai}, converting each map into an atomic structure, a process known as atomic model building, is complicated by the resolution of the heterogeneous volumes (maps). Existing learning-based model building algorithms \cite{jamali2024automated} often produce incomplete structures because they are trained on relatively high-resolution cryo-EM maps. Emerging methods \cite{Fadini2026, raghu2025multiscale} use AlphaFold or Boltz-2 \cite{passaro2025boltz} as priors to guide model building but face sub-optimal runtime and memory usage and output structures with inaccurate stereochemistry.

Often, the atomic conformations produced by model building algorithms are deposited separately in the PDB, where they may subsequently be utilized to train atomic ensemble prediction models such as BioEmu \cite{lewis2025scalable} or Boltz-2 \cite{passaro2025boltz}. However, the difficulty of model building for cryo-EM ensembles means that the PDB lacks data on protein dynamics that currently exist in public experimental datasets. Our key insight in this work is to combine the two stages of model building and structure training into a single procedure that supervises Boltz-2 directly on raw cryo-EM density maps. Our method, CryoSampler, fine-tunes Boltz-2 with an ensemble head that trains on the entire cryo-EM map ensemble simultaneously.

We primarily validate our technique at the overfitting task of fine-tuning Boltz-2 to an ensemble of maps derived from a single cryo-EM experiment, corresponding to a single protein system undergoing motion. When trained in this manner, CryoSampler performs \textit{atomic ensemble model building}, converting an ensemble of maps into an ensemble of atomic conformations. By training our model via end-to-end 3D reconstruction-based optimization, we achieve state-of-the-art performance in terms of map--model fit and stereochemistry.

Future atomic ensemble prediction methods would greatly benefit from training on cryo-EM datasets, which are difficult to capture with other techniques. In principle, it is possible to extend our training strategy and fine-tune Boltz-2 on all publicly-available cryo-EM ensemble data in order to generalize across different sequences. However, due to the sparsity of public cryo-EM ensemble data, we do not demonstrate out-of-domain generalization in this work. Rather, we offer initial evidence of in-domain generalization within the family of transient receptor potential (TRP) proteins, showing that it is possible to transfer learned conformational features from one TRP channel to another with higher accuracy than general-purpose ensemble prediction techniques.

\begin{figure}
    \centering
    \includegraphics[width=0.8\textwidth]{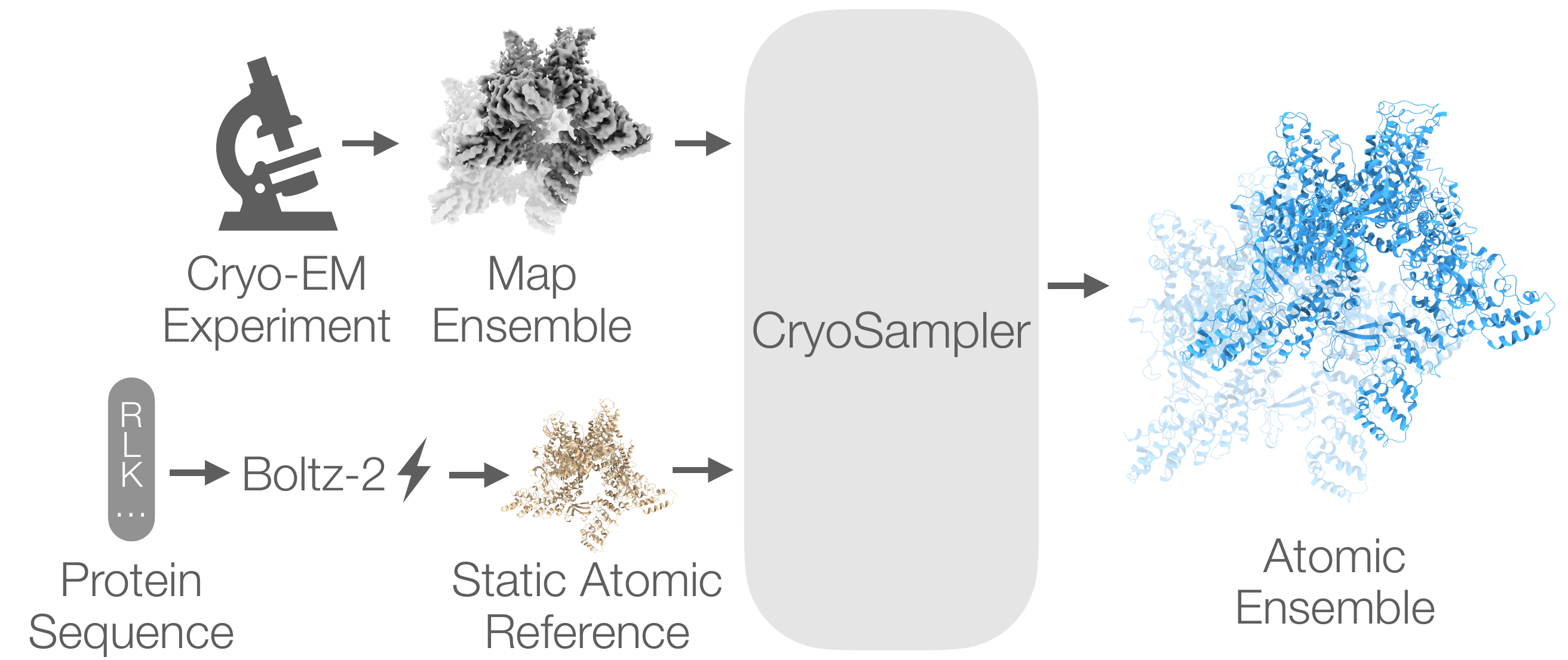}
    \captionof{figure}{Our method, CryoSampler, fine-tunes Boltz-2 \cite{passaro2025boltz}, a state-of-the-art static structure predictor, to generate atomic conformational ensembles via raw supervision on cryo-EM map ensembles. We train a latent diffusion-based head on top of the frozen output of Boltz-2 to output atomic conformations with high agreement to the input maps as well as valid stereochemistry.}
    \label{fig:teaser}
\end{figure}

\section{Prior Work}
\label{sec:prior_work}

Previous work has typically trained atomic ensemble prediction models through a two-stage process: experimental maps from cryo-EM (and occasionally, X-ray crystallography) are converted into atomic structures, which are subsequently incorporated into the training data of ensemble predictors. Because building atomic models into cryo-EM ensembles is difficult due to resolution issues, practitioners will often deposit a single ensemble-averaged structure into the PDB rather than the full structural ensemble, so models downstream remain blind to this rich source of dynamics. In the following sections, we describe the limitations of the two-stage process and show how our unified approach of supervising pre-trained structure predictors on raw cryo-EM maps resolves some of these challenges.

\subsection{Model Building}

Building atomic models into cryo-EM maps has traditionally been a manual process involving 3D graphics programs such as Coot \cite{Emsley:ba5144} and ISOLDE \cite{Croll:ic5101}. ModelAngelo \cite{jamali2024automated} was one of the first fully-automated workflows to achieve competitive performance using a machine learning algorithm trained on thousands of cryo-EM maps. While ModelAngelo surpasses human capabilities and operates completely de novo (without an initial structure), it often produces incomplete models in ambiguous or low-resolution regions.

More recent methods have incorporated AlphaFold as a prior to improve model completeness. ROCKET \cite{Fadini2026} performs fitting by optimizing the multiple sequence alignment (MSA) embeddings of AlphaFold 2. CryoBoltz \cite{raghu2025multiscale} guides the sampling procedure of Boltz-1 \cite{wohlwend2025boltz} via diffusion posterior sampling in order to fit a target cryo-EM map. Chen \cite{chen2025building} describes a complementary fitting approach that jointly optimizes 2D map--model correlation and atomic geometry using differentiable approximations of MolProbity metrics. While these approaches leverage pre-trained models to enhance model completion, they often face runtime issues or output structures with invalid stereochemistry.

Unlike existing techniques that only utilize the inference capabilities of pre-trained models, CryoSampler adopts a training-time approach that fine-tunes Boltz-2 to generate an atomic conformational ensemble that best describes an input ensemble of cryo-EM maps. Our unique training strategy holds the potential to allow future protein foundation models to supervise directly on raw cryo-EM ensembles, implicitly performing model building during training rather than as a pre-processing step.

\subsection{Ensemble Prediction}

Learning-based methods that predict atomic ensembles purely from sequence can be broadly divided into those that operate at training time and those that modify pre-trained static prediction models such as AlphaFold \cite{jumper2021highly, abramson2024accurate} or Boltz \cite{wohlwend2025boltz, passaro2025boltz} at inference time \cite{sledzieski2026landscape}. Inference-time approaches alter various components of the base model, such as the multiple sequence alignment input \cite{wayment2024predicting, monteiro2024high}, pair representation \cite {suzuki2026steering}, or diffusion model \cite{richman2025unlocking}, in order to generate diverse conformations. These methods are constrained by the relatively static nature of PDB-derived training data, as they only operate within the weight space of models trained on PDB (or self-distilled) data.

On the other hand, models such as BioEmu \cite{lewis2025scalable} and AlphaFlow \cite{jing2024alphafold} enrich the training set with simulation data from molecular dynamics (MD). Although MD offers data on conformational variability that may not exist in the PDB, it remains computationally challenging to simulate the dynamics of larger protein complexes in a physically-accurate manner. Cryo-EM is well-suited to studying the motions of these complexes, and because our method supervises directly on cryo-EM maps, it has the potential to scale to systems that elude existing ensemble generation techniques.
\section{Methods}
\label{sec:methods}

\begin{figure}[t]
    \centering
    \includegraphics[width=\textwidth]{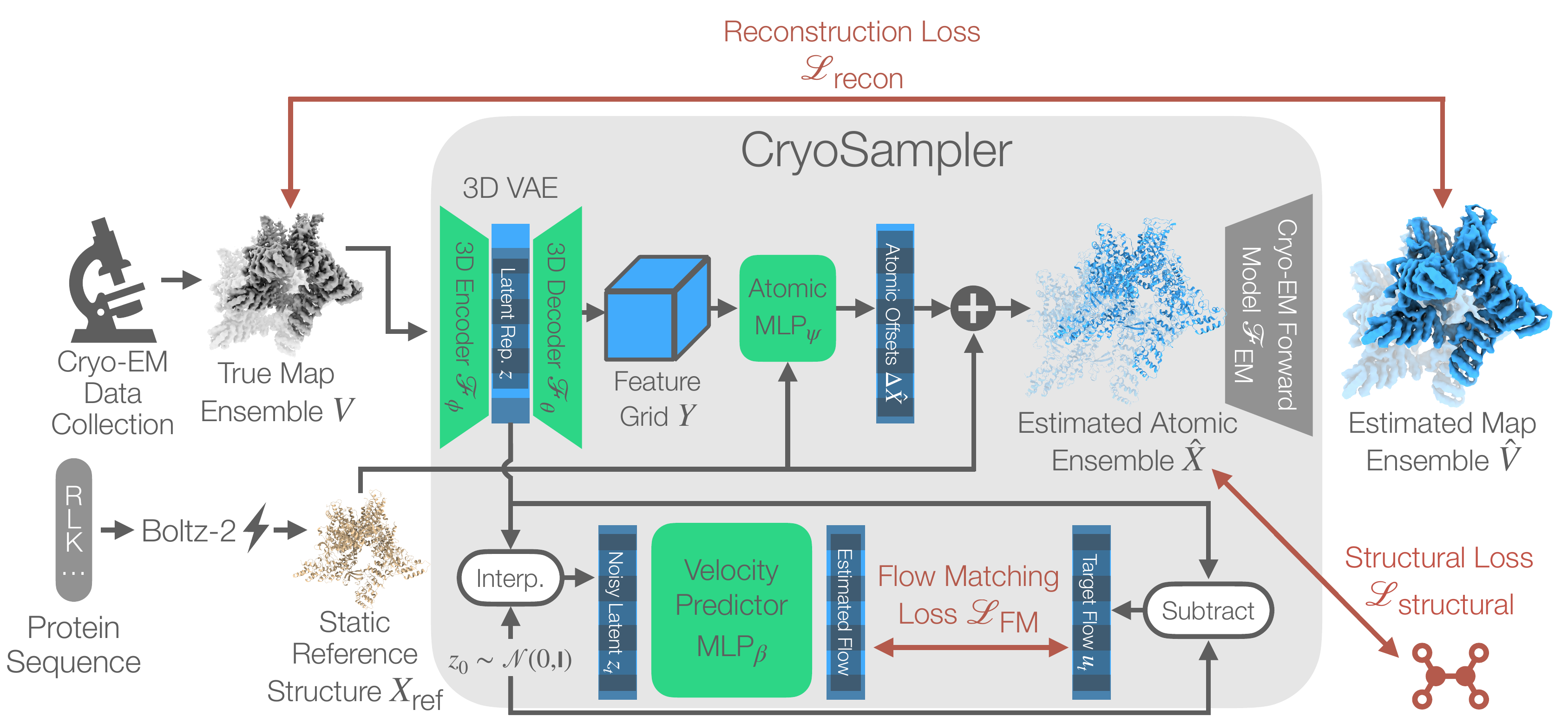}
    \caption{CryoSampler is a latent diffusion model that is trained in two stages: in the first, we optimize a 3D variational autoencoder (VAE) to perform atomic model building on an input ensemble of experimental cryo-EM maps. The autoencoder learns how to deform a static reference structure from Boltz-2, $ X_{\text{ref}} $, in order to produce an ensemble of atomic conformations that when rendered differentiably using the physics of the cryo-EM forward model, produces an estimated map ensemble $ \hat{V} $ that is supervised end-to-end against the input maps. In the second stage of training, the VAE is frozen, and we train a latent diffusion model on the embeddings from the first stage via a flow matching objective.}
    \label{fig:training_diagram}
    \vspace{-10pt}
\end{figure}

CryoSampler is a latent diffusion model (LDM) that performs model building on an input ensemble of cryo-EM maps that captures the various conformations of a protein. As shown in Figure \ref{fig:training_diagram}, the model learns a distribution over the input maps via 3D volumetric reconstruction. The underlying distribution that we wish to model is that of protein conformations $ X $ expressed in terms of 3D atomic coordinates. Unlike prior methods that train directly on these coordinates, we do not assume that we have access to samples from the atomic distribution. Rather, our samples are represented in the space of cryo-EM maps $ \mathcal{F}_{\text{EM}}(X) $, where $ \mathcal{F}_{\text{EM}} $ represents the cryo-EM forward model that converts atomic coordinates into 3D density fields.

While $ \mathcal{F}_{\text{EM}} $ has an analytical form that facilitates rendering maps from atomic coordinates, the reverse process represents the model building problem, which is underdetermined. Our model conducts model building for an input ensemble of maps via end-to-end supervision on volumetric reconstruction. Training is carried out in two stages: in the first, we optimize a 3D variational autoencoder (VAE) to learn the distribution of atomic offsets needed to deform a static atomic structure output by Boltz-2 \cite{passaro2025boltz} in order to match the input maps in a self-supervised manner. In the second stage, the VAE is frozen, and we train a flow matching model on the latent embeddings from the previous stage. At inference time, one can sample atomic ensembles for new protein sequences not seen during training. Our model's pipeline is described in more detail in the following sections.

\subsection{Training}

\subsubsection{Model Building With a 3D Variational Autoencoder}

During the first stage of training, a 3D spatial variational autoencoder (VAE) takes an ensemble of cryo-EM maps as input and learns a distribution of deformations on top of the static atomic structure output by Boltz-2. The weights of Boltz-2, as well as its structural output, are frozen during our training procedure. At a high level, each input map of resolution $ R^3 $ is encoded into a compressed latent representation, which is then decoded into a feature grid at the original map's resolution that represents a hybrid neural field of the 3D atomic offsets to apply to the static reference structure $ X_{\text{ref}} $. This field is queried at the coordinates of $ X_{\text{ref}} $, producing 3D offsets that are added back to the reference coordinates, yielding an estimated atomic model $ \hat{X} $. The atomic model is then rendered into an estimated map $ \hat{V} $ using a differentiable physics-based forward model of the cryo-EM capture process \cite{tang2007eman2}, and we compute an end-to-end reconstruction loss on the input and output maps to optimize the weights of the VAE.

The VAE's encoder $ \mathcal{F}_\phi $ compresses the input map $ V $ into a lower-dimensional continuous latent grid distribution $ \mathcal{N}(\mu, \sigma^2) $ at 1/8th the spatial resolution using ConvNeXt-style blocks for memory efficiency. We sample a latent embedding $ z $ using the standard reparameterization trick:

\begin{equation}
z = \mu_\phi(V) + \exp(0.5 \cdot \log\sigma^2_\phi(V)) \odot \epsilon \quad \text{where} \quad \epsilon \sim \mathcal{N}(0, \mathbf{I}),
\end{equation}

where $ \mu_\phi(V), \sigma^2_\phi(V) $ represent the mean and variance output of the encoder. The decoder $ \mathcal{F}_\theta $ expands this latent representation into a 3D feature grid $ Y = \mathcal{F}_\theta(z) $ with 64 channels that matches the original volume's spatial dimensions.  This feature grid is decoded into a discrete set of coordinate offsets by first applying trilinear interpolation via grid sampling at the Boltz-2 reference coordinates. The interpolated per-atom features are then passed through a lightweight multi-layer perceptron (MLP) to predict the per-atom offsets, which are added to the Boltz-2 coordinates to produce the estimated atomic model $ \hat{X} $. Lastly, the coordinates are differentiably rendered into an estimated map $ \hat{V} $ using the cryo-EM forward model $ \mathcal{F}_{\text{EM}} $. Put together, the decoder can be seen as a type of hybrid neural field, represented using the following equations:

\begin{equation}
\hat{V} = \mathcal{F}_{\text{EM}} \left( X_{\text{ref}} + \text{MLP}_\psi \left( X_{\text{ref}}, \text{GridSample}(Y, X_{\text{ref}}) \right) \right)
\end{equation}

The VAE is optimized end-to-end via a reconstruction loss $ \mathcal{L}_{\text{recon}} $ and a KL-divergence penalty $ D_{\text{KL}} $ on the latent embeddings $ z $. The reconstruction loss is computed as the inverse of the normalized cross-correlation (NCC) between the input and output volumes $ V, \hat{V} $:

\begin{equation}
\mathcal{L}_{\text{recon}} = 1 - \text{NCC}(\hat{V}, V),
\end{equation}

\begin{equation}
\text{NCC}(\hat{V}, V) = \frac{\sum_{i=1}^N (\hat{V}_i - \hat{\mu}_{\text{vol}})(V_i - \mu_{\text{vol}})}{\left(\sqrt{\sum_{i=1}^N (\hat{V}_i - \hat{\mu}_{\text{vol}})^2 + \epsilon}\right) \left(\sqrt{\sum_{i=1}^N (V_i - \mu_{\text{vol}})^2 + \epsilon}\right) + \epsilon},
\end{equation}

where $ \mu_{\text{vol}}, \hat{\mu}_{\text{vol}} $ represent the voxel-wise means of the input and output map ensembles, respectively, $ i $ indexes the volume in each ensemble, and $ \epsilon = 10^{-8} $ is used to prevent numerical instability.

We impose a structural loss $ \mathcal{L}_{\text{structural}} $ to regularize the prediction of the atomic ensemble $ \hat{X} $, consisting of differentiable versions of PDB validation metrics first introduced in \cite{chen2025building} and a novel soft rigid body deformation loss operating purely on the backbone atoms of $ \hat{X} $. The PDB validation losses $ \mathcal{L}_{\text{MolProb}} $ regularize the stereochemistry of the estimated atomic structures, minimizing clashes between the atoms and keeping the backbone dihedral angles and side-chain torsion angles within physically valid ranges.

The backbone geometry loss $ \mathcal{L}_{\text{backbone}} $ aims to preserve secondary structural features in the reference structure, such as alpha helices, by constraining local geometry across adjacent residues. The backbone loss penalizes deviations from the pairwise distances of the reference backbone atoms, effectively acting as a soft rigid body constraint for the deformation field, and is formulated as:

\begin{equation}
\mathcal{L}_{\text{backbone}} = \frac{1}{M \cdot |P_B|} \sum_{m=1}^M \sum_{(i,j) \in P_B} \left( || \hat{X}_i^{(m)} - \hat{X}_j^{(m)} ||_2 - || X_{\text{ref},i}^{(m)} - X_{\text{ref},j}^{(m)} ||_2 \right)^2
\end{equation}

Here, $ M = N_{\text{res}} - W + 1 $ is the total number of sliding windows, $ \hat{X}^{(m)} $ and $ X_{\text{ref}}^{(m)} $ denote the coordinates of the backbone atoms in window $ m $, and $ P_B $ is the set of all unique pairwise combinations of the 20 backbone atoms within that window.

The complete optimization objective for training the VAE, including both the volumetric and atomic losses, can be written as:

\begin{equation}
\mathcal{L}_{\text{VAE}} = \mathcal{L}_{\text{recon}}(V, \hat{V}) + \lambda_{\text{KL}} D_{\text{KL}}(q(z|V) || p(z)) + \mathcal{L}_{\text{structural}},
\end{equation}

where $ p(z) $ is the standard normal. The structural loss consists of the previous backbone loss and a stereochemical loss $ \mathcal{L}_{\text{MolProb}} $ introduced in \cite{chen2025building}, which corresponds to differentiable approximations of standard MolProbity metrics that measure the validity of each structure's dihedral angles (Ramachandran outliers), side chain angles (rotamer outliers), and inter-atomic clashes. The full structural loss is then:

\begin{equation}
\mathcal{L}_{\text{structural}} = \lambda_M \mathcal{L}_{\text{MolProb}} + \lambda_B \mathcal{L}_{\text{backbone}}
\end{equation}

Empirically, we find that that setting $ \lambda_{\text{KL}} = 10^{-4}, \lambda_M = 10^{-6}, \lambda_B = 10^2 $ yields atomic models with high map--model fit and stereochemically accurate geometry.

\subsubsection{Latent Diffusion Model}

We train a diffusion model on the latent embeddings produced by the VAE in the previous section. The weights of the VAE are frozen, and the mean latent codes $ \mu $ for all conformational states in the training set of cryo-EM maps are extracted. This stage of training employs a time-conditioned $ \text{MLP}_\beta $ that operates on the compressed $V/8$ latent grid, optimized with a continuous-time flow matching objective. The VAE latent grid of shape $(C \times (V/8)^3)$ is flattened into a single vector, which is concatenated with a sinusoidal embedding of the continuous timestep $t$ before being passed through four fully-connected layers of width $d$ with SiLU activations and a linear readout. The model learns a vector field that smoothly interpolates between a standard normal prior distribution $z_0 \sim \mathcal{N}(0, \mathbf{I})$ and the target distribution of latent codes $z_1 = \mu$. The linear interpolation between these two distributions can be written as:

\begin{equation}
z_t = t z_1 + (1 - t) z_0, \quad t \in [0, 1]
\end{equation}

The target vector field is the straight line vector from the pure noise sample to the latent sample:

\begin{equation}
u_t = z_1 - z_0
\end{equation}

During optimization, the flow matching objective aims to optimize the weights of the Diffusion Transformer $ \mathcal{F}_{\psi} $ to match the target vector field, over all possible samples $ z_0, z_1 $ and interpolation times $ t $:

\begin{equation}
\mathcal{L}_{\text{FM}} = \mathbb{E}_{t, z_0, z_1} \left[ || \mathcal{F}_{\psi}(z_t, t) - u_t ||^2 \right]
\end{equation}

\subsection{Inference: In-Family Ensemble Prediction}

After training the latent diffusion model on cryo-EM map ensembles, it is able to generate atomic ensembles for unseen proteins at inference time, bypassing the need for experimental density maps. This capability is validated on new proteins that lie in the same family as the training set and exhibit similar motions, displaying in-domain generalization. During inference, the sampling process starts with pure Gaussian noise samples in the latent space:

\begin{equation}
z_0 \sim \mathcal{N}(0, \mathbf{I})
\end{equation}

The latent model $ \text{MLP}_\beta $ is evaluated over 50 discrete integration steps to yield a generated ensemble of latent grids $ z_1 $. The sampled latent grids are then passed through the VAE decoder to produce the 3D feature grid. The deformation process then proceeds in the same manner as in training, except using the coordinates of the target protein $ \tilde{X}_{\text{ref}} $ (generated by Boltz-2) for sampling the feature grid. The trained atomic MLP outputs the offsets used to deform the reference structure of the target protein. The analytical form of the generated coordinates is then:

\begin{equation}
X_{\text{final}} = \text{MLP}_{\psi}\left( \tilde{X}_{\text{ref}}, \text{GridSample}(\mathcal{F}_{\theta}(z_1), \tilde{X}_{\text{ref}}) \right)
\end{equation}
\section{Results}
\label{sec:results}

We evaluate our method with two capabilities, building atomic models into cryo-EM maps, and generating conformational ensembles of related proteins in the same family, and find that it exceeds the performance of existing methods in both areas. We note that because our experiments involve real data from cryo-EM experiments, we do not have access to ``ground truth" atomic structures for the cryo-EM maps. Thus, in the absence of ground truth, we use the same metrics that the Protein Data Bank \cite{berman2000protein} uses to evaluate structure depositions: map--model fit and model geometry. These metrics are computed using the software package Phenix \cite{liebschner2019macromolecular} and respectively measure the cross-correlation between the input map and the estimated atomic model, as well as the physical validity of the atomic model's dihedral angles, side-chain torsion angles, and inter-atomic distances (clashes).

\subsection{Model Building for Experimental Cryo-EM Maps}

\begin{figure}[t]
    \centering
    \includegraphics[width=0.9\textwidth]{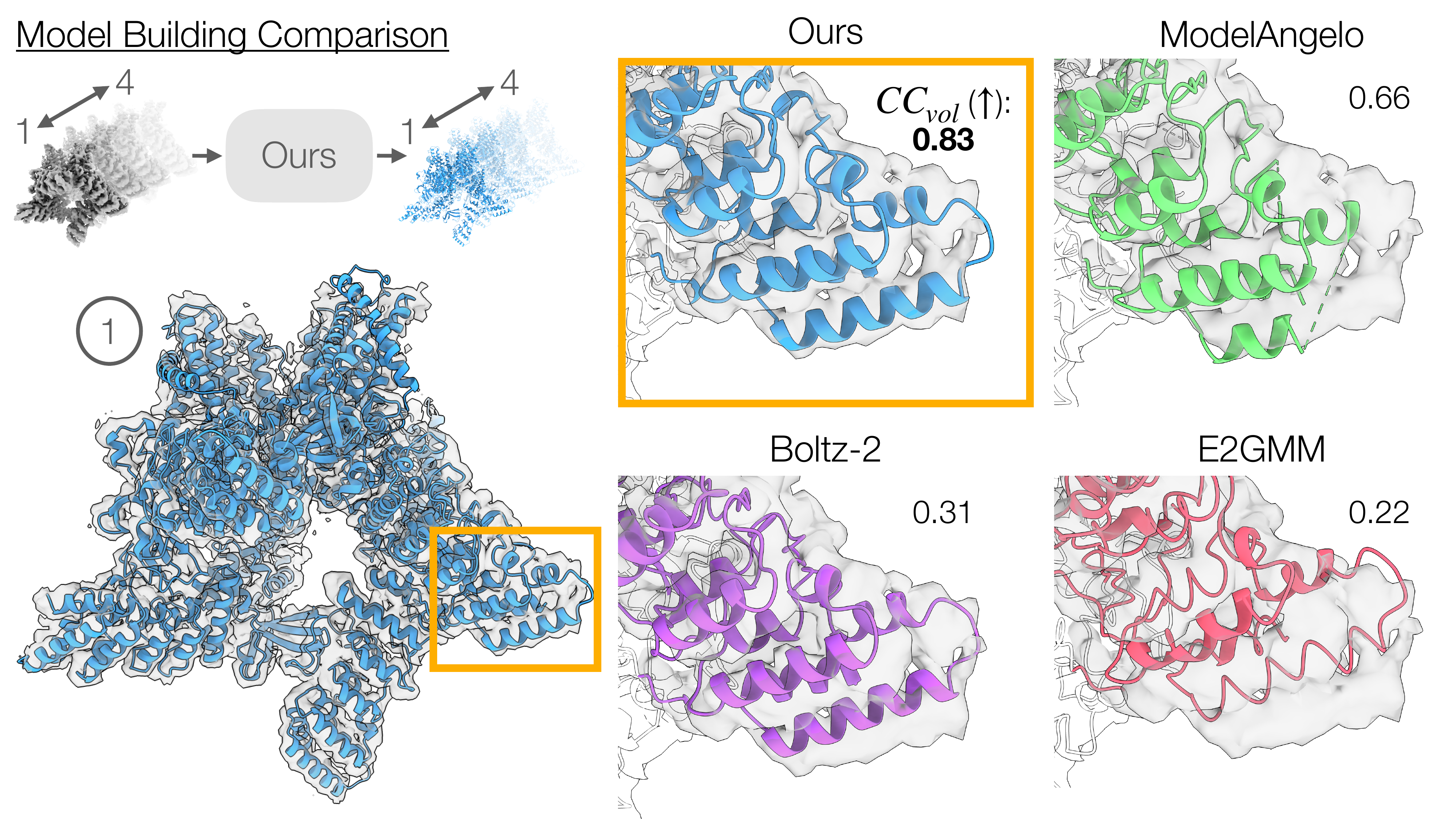}
    \caption{Model building performance comparison visualized the TRPV3 channel protein \cite{shimada2020structure} processed via 3D Variability Analysis \cite{punjani20213d}. The experimental cryo-EM map is shaded in transparent gray. Our method is run on four input maps jointly, and we visualize the results on the first of these maps. Among all other methods, ours achieves the best map--model fit against the observed map, as shown visually in the zoom inset and quantitatively as assessed by the cross-correlation metric $ CC_{volume} $ in Phenix \cite{liebschner2019macromolecular}. ModelAngelo achieves the second highest map--model fit for this map, but its modeled structure is incomplete and contains missing regions, as shown in the rightmost extent of the inset map.}
    \label{fig:model_building_comparison}
\end{figure}

We evaluate the model building accuracy of the VAE on a variety of biological systems captured with cryo-EM experiments: the transient receptor potential vanilloid subfamily member 3 (TRPV3) protein expressing a ``breathing” motion \cite{shimada2020structure}, the integrin $ \alpha V \beta 8 $ with a flexible ``tail” \cite{campbell2020cryo}, the neurokinin-1 G protein-coupled receptor (GPCR) \cite{harris2022selective}, and human P-glycoprotein undergoing drug transport \cite{culbertson2025cryo}. Each ensemble contains four maps; we reconstruct the first three datasets with 3D Variability Analysis \cite{punjani20213d} in CryoSPARC, while the latter is retrieved from the PDB.

For the TRPV3, integrin, and GPCR proteins, the VAE conducts model building using all the maps jointly, using a single sample from Boltz-2 as the static reference structure, whereas for the P-glycoprotein, model building is carried out on each input map individually. As shown qualitatively in Figure \ref{fig:model_building_comparison}, our method outperforms prior work and determines atomic models with the best fit to the data, as measured by $ CC_{volume} $. The native output of Boltz-2 and E2GMM, which output complete structures, have lower map--model fit. ModelAngelo achieves the second highest fit on TRPV3, but as seen in the zoom inset of the figure, has incomplete regions in certain areas of the map. CryoBoltz runs out of memory on this protein complex, which contains 2,556 residues.

As seen in Table \ref{table:model_building_metrics_full}, our method achieves the highest map--model fit among all other methods when considering all forms of cross-correlation metrics available in Phenix. It also produces atomic models with accurate stereochemistry, achieving competitive MolProbity scores, which combine Ramachandran outliers, side chain outliers, and clash scores into a single metric. The full geometry metrics are reported in the supplement. With these results, CryoSampler establishes state-of-the-art performance in model building for map ensembles.

\subsection{Ensemble Prediction for TRPV5 Channel Protein}

\begin{figure}[t]
    \centering
    \includegraphics[width=\textwidth]{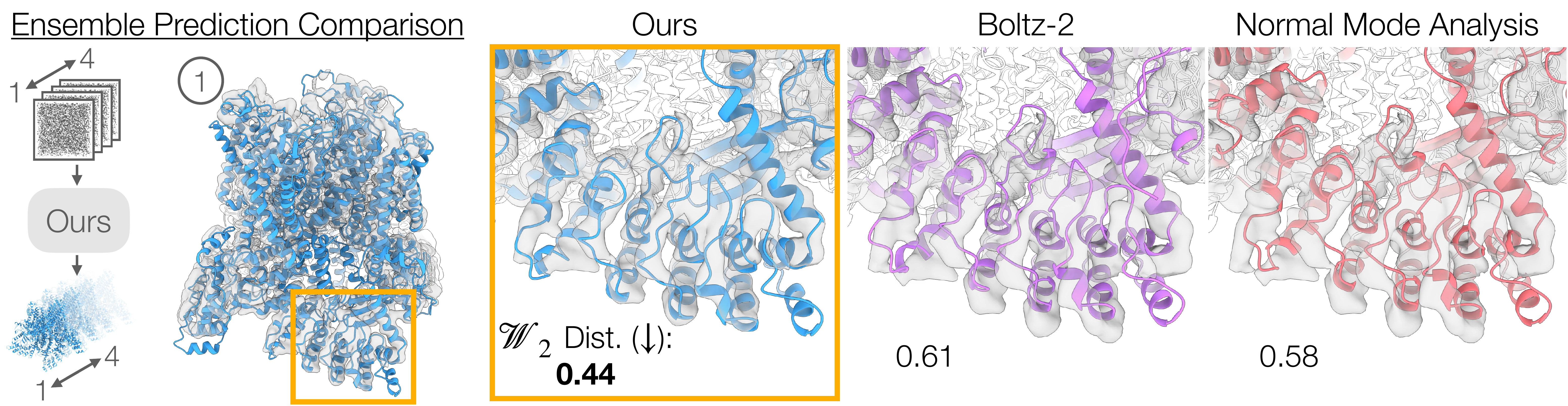}
    \caption{Ensemble prediction performance comparison. In this experiment, we train on an ensemble of four maps of the TRPV3 channel protein \cite{shimada2020structure}, and evaluate each method's ability to predict the ensemble for a distinct TRPV5 protein \cite{dang2019structural}, assessed against a held-out validation set of four cryo-EM maps derived from 3D Variability Analysis \cite{punjani20213d}. We visualize one of the ground truth TRPV5 maps in transparent gray, as well as the closest generated atomic structure for each method to this map. As shown in the zoom insets, our method produces an atomic conformation that fits the cryo-EM map most faithfully. Quantitatively, we achieve the lowest Wasserstein ($ \mathcal{W}_2 $) distance, which measures ensemble accuracy pairwise across all generated atomic conformations and true maps.}
    \label{fig:ensemble_prediction_comparison}
\end{figure}

\begin{table}[h]
  \caption{Performance comparison on generating the ensemble for TRPV5, with no cryo-EM data input to the model. We train our model on 4 states of the TRPV3 channel protein, extracted via 3D Variability Analysis \cite{punjani20213d}, and measure the ability of all methods to produce atomic structures that are faithful to a held-out validation set of maps of TRPV5. To evaluate ensemble prediction accuracy, we compute pairwise map--model fit in terms of $ CC_{volume} $ between all generated atomic structures and all cryo-EM maps, subsequently computing the precision, recall, and Wasserstein ($ \mathcal{W}_2 $) distance using $ \sqrt{1 - CC_{volume}} $ as the underlying metric. Our method achieves the best ensemble prediction accuracy compared to all prior techniques.}
  \label{table:ensemble_metrics}
  \centering
  \begin{tabular}{p{1.9cm}cccc}
    \toprule
    & Boltz-2 \cite{passaro2025boltz} & NMA \cite{bakan2011prody} & \textbf{Ours} \\
    \midrule
    Precision ($\uparrow$) & 0.41 & 0.43 & \textbf{0.58} \\
    Recall ($\uparrow$)    & 0.43 & 0.43 & \textbf{0.59} \\
    $\mathcal{W}_2$ Dist. ($\downarrow$) & 0.61 & 0.58 & \textbf{0.44} \\
    \bottomrule
  \end{tabular}
\end{table}

We show that when trained on an ensemble of cryo-EM maps, our latent diffusion model can sample atomic conformations for new proteins that reside within the same family and exhibit similar conformational variability. To evaluate this capability, we process two single particle cryo-EM datasets of distinct TRP channel proteins, TRPV3 \cite{shimada2020structure} and TRPV5 \cite{dang2019structural}, and use 3DVA \cite{punjani20213d} to produce map ensembles for both. Further details regarding our reconstruction procedure can be found in the supplement. We then identify one principal component corresponding to a breathing motion shared between the two proteins, and select four density maps for each, spaced evenly along the principal component. CryoSampler is trained on the TRPV3 maps, and at inference time, we evaluate its capability to produce atomic ensembles of TRPV5 that match the held-out validation set of TRPV5 maps.

We sample 16 conformations of TRPV5 from our method and each of the baseline methods. AlphaFlow \cite{jing2024alphafold} and BioEmu \cite{lewis2025scalable} are excluded from evaluation because they only sample conformations for monomeric proteins containing single chains, and cannot handle the tetrameric (4-chain) structure of TRPV5. Boltz-sample \cite{suzuki2026steering} runs out of memory when computing atomic conformations for TRPV5, which contains 2,920 residues. These atomic structures are then aligned with and compared against the experimental TRPV5 maps using the $ CC_{volume} $ metric, from which we compute precision, recall, and Wasserstein ($ W_2 $) distance at the ensemble level. As shown in Figure \ref{fig:ensemble_prediction_comparison} and Table \ref{table:ensemble_metrics}, our method samples conformations that exhibit best fit to the experimental data. Thus, we show that CryoSampler is able to perform zero-shot ensemble prediction for TRPV5 via conformational features learned from TRPV3, outperforming general-purpose ensemble prediction methods at this task.

\begin{table}[h!]
  \caption{Model building performance comparison on four systems: TRPV3 \cite{shimada2020structure}, P-glycoprotein \cite{culbertson2025cryo}, integrin $ \alpha V \beta 8 $ \cite{campbell2020cryo}, and the neurokinin-1 GPCR \cite{harris2022selective}. The methods are assessed based on map--model fit and model geometry in Phenix \cite{liebschner2019macromolecular}, which is standard practice in the field. Our method achieves the highest map--model fit for nearly every cross-correlation metric across every system studied when compared to prior work. The quality of each method's model geometry is assessed via the MolProbity score, which combines information about Ramachandran and side chain outliers as well as atomic clashes. We maintain competitive quality for model geometry, achieving best or second-best performance across every system except the P-glycoprotein. For each table entry, we report standard deviation error bars across $n=3$ replicates. Bold values indicate the best-performing method(s), and values within one standard deviation of the best are also bolded where ties occur. A full version of this table, including detailed model geometry metrics and method runtimes, can be found in the supplement. *ModelAngelo only built models for 3 out of 4 maps for these systems.}
  \label{table:model_building_metrics_compact}
  \centering
  \small
  \begin{tabular}{llccccc}
    \toprule
    & Metric & Boltz-2 \cite{passaro2025boltz} & CryoBoltz \cite{raghu2025multiscale} & E2GMM \cite{chen2025building} & ModelAngelo \cite{jamali2024automated} & Ours \\
    \midrule
    \multirow{7}{*}{\rotatebox[origin=c]{90}{TRPV3}} & \textit{Map--Model Fit} &  &  &  &  &  \\
    & $CC_{volume}$ ($\uparrow$) & 0.31$\pm$0.03 & --- & 0.25$\pm$0.02 & \underline{0.79$\pm$0.08} & \textbf{0.84$\pm$0.02} \\
    & $CC_{mask}$ ($\uparrow$) & 0.30$\pm$0.03 & --- & 0.23$\pm$0.02 & \underline{0.79$\pm$0.07} & \textbf{0.84$\pm$0.02} \\
    & $CC_{peaks}$ ($\uparrow$) & 0.29$\pm$0.03 & --- & 0.21$\pm$0.02 & \underline{0.71$\pm$0.08} & \textbf{0.79$\pm$0.02} \\
    & $CC_{box}$ ($\uparrow$) & 0.62$\pm$0.02 & --- & 0.55$\pm$0.01 & \underline{0.85$\pm$0.04} & \textbf{0.89$\pm$0.01} \\
    \cmidrule{2-7}
    & \textit{Model Geometry} &  &  &  &  &  \\
    & MP Score ($\downarrow$) & 1.83$\pm$0.04 & --- & \textbf{1.57$\pm$0.07} & 3.00$\pm$0.16 & \underline{1.64$\pm$0.04} \\
    \midrule
    \multirow{7}{*}{\rotatebox[origin=c]{90}{P-glycoprotein}} & \textit{Map--Model Fit} &  &  &  &  &  \\
    & $CC_{volume}$ ($\uparrow$) & 0.39$\pm$0.04 & \underline{0.73$\pm$0.03} & 0.26$\pm$0.02 & 0.59$\pm$0.15$^*$ & \textbf{0.77$\pm$0.02} \\
    & $CC_{mask}$ ($\uparrow$) & 0.40$\pm$0.03 & \underline{0.74$\pm$0.03} & 0.26$\pm$0.03 & 0.59$\pm$0.14 & \textbf{0.77$\pm$0.02} \\
    & $CC_{peaks}$ ($\uparrow$) & 0.53$\pm$0.12 & \underline{0.70$\pm$0.03} & 0.44$\pm$0.13 & -0.11$\pm$0.16 & \textbf{0.75$\pm$0.05} \\
    & $CC_{box}$ ($\uparrow$) & \underline{0.62$\pm$0.07} & \textbf{0.85$\pm$0.02} & 0.54$\pm$0.08 & 0.33$\pm$0.09 & \textbf{0.85$\pm$0.02} \\
    \cmidrule{2-7}
    & \textit{Model Geometry} &  &  &  &  &  \\
    & MP Score ($\downarrow$) & \underline{1.82$\pm$0.06} & 2.98$\pm$0.41 & \textbf{1.60$\pm$0.11} & 2.89$\pm$1.07 & 2.67$\pm$0.26 \\
    \midrule
    \multirow{7}{*}{\rotatebox[origin=c]{90}{Integrin}} & \textit{Map--Model Fit} &  &  &  &  &  \\
    & $CC_{volume}$ ($\uparrow$) & \underline{0.62$\pm$0.07} & 0.38$\pm$0.13 & 0.35$\pm$0.05 & \underline{0.58$\pm$0.03}$^*$ & \textbf{0.68$\pm$0.05} \\
    & $CC_{mask}$ ($\uparrow$) & \underline{0.61$\pm$0.06} & 0.39$\pm$0.12 & 0.29$\pm$0.06 & \underline{0.57$\pm$0.02} & \textbf{0.68$\pm$0.05} \\
    & $CC_{peaks}$ ($\uparrow$) & \underline{0.75$\pm$0.04} & \underline{0.72$\pm$0.05} & 0.59$\pm$0.04 & -0.22$\pm$0.03 & \textbf{0.81$\pm$0.03} \\
    & $CC_{box}$ ($\uparrow$) & \underline{0.75$\pm$0.04} & \underline{0.72$\pm$0.05} & 0.59$\pm$0.04 & 0.28$\pm$0.02 & \textbf{0.81$\pm$0.03} \\
    \cmidrule{2-7}
    & \textit{Model Geometry} &  &  &  &  &  \\
    & MP Score ($\downarrow$) & \textbf{1.80$\pm$0.07} & 3.56$\pm$0.08 & \underline{1.88$\pm$0.09} & 4.03$\pm$0.13 & \textbf{1.80$\pm$0.07} \\
    \midrule
    \multirow{7}{*}{\rotatebox[origin=c]{90}{NK-1 GPCR}} & \textit{Map--Model Fit} &  &  &  &  &  \\
    & $CC_{volume}$ ($\uparrow$) & 0.55$\pm$0.06 & \textbf{0.73$\pm$0.05} & \textbf{0.75$\pm$0.04} & \underline{0.69$\pm$0.06} & \textbf{0.73$\pm$0.04} \\
    & $CC_{mask}$ ($\uparrow$) & 0.55$\pm$0.06 & \textbf{0.72$\pm$0.05} & \textbf{0.75$\pm$0.04} & \underline{0.70$\pm$0.06} & \textbf{0.72$\pm$0.04} \\
    & $CC_{peaks}$ ($\uparrow$) & \underline{0.55$\pm$0.10} & \textbf{0.70$\pm$0.05} & \textbf{0.74$\pm$0.07} & 0.29$\pm$0.30 & \textbf{0.71$\pm$0.08} \\
    & $CC_{box}$ ($\uparrow$) & 0.75$\pm$0.04 & \textbf{0.84$\pm$0.03} & \textbf{0.85$\pm$0.02} & 0.59$\pm$0.22 & \underline{0.83$\pm$0.03} \\
    \cmidrule{2-7}
    & \textit{Model Geometry} &  &  &  &  &  \\
    & MP Score ($\downarrow$) & \textbf{1.87$\pm$0.02} & \underline{3.01$\pm$0.05} & \textbf{1.75$\pm$0.20} & 3.61$\pm$0.20 & \textbf{1.93$\pm$0.03} \\
    \bottomrule
  \end{tabular}
\end{table}

\section{Discussion}
\label{sec:discussion}

In this work, we introduce a principle for training diffusion models of protein ensembles directly on cryo-EM density maps, demonstrating that a unified model can both perform atomic model building and ensemble generation. Our technique, CryoSampler, shows that conducting VAE-based volumetric and structural optimization using Boltz-2 \cite{passaro2025boltz} as a static trunk achieves higher map--model fit and competitive model geometry compared to prior work. Once trained, our model is capable of generating atomic structures for unseen proteins in the same class that exhibit similar modes of flexibility, demonstrated on two proteins in the TRP channel family.

We primarily validate CryoSampler's ability to build atomic models from map ensembles. While our model is capable of zero-shot ensemble generation for novel sequences, we only provide preliminary evidence of this capability across two proteins within the same TRP channel family (in-domain generalization). Demonstrating out-of-domain generalization would require processing a greater variety of cryo-EM datasets across different families, and potentially conditioning the generation process on sequence-derived features similar to AlphaFold 3 \cite{abramson2024accurate}. Showing generalization across diverse protein families is a promising avenue for future work, and our method lays the building blocks for training larger foundation models directly on raw cryo-EM map data.

In addition to establishing proof of generalization, future work could explore the idea of training generative models on other types of raw observational data. While traditional ensemble prediction techniques operate in the space of atomic coordinates, we show that treating cryo-EM map data as the distribution being modeled can achieve superior results in the specific case of in-family conformational feature transfer. Further research can be carried out on extending this principle to various modalities of biological data, such as crystallographic structure factors and nuclear magnetic resonance spectra, to fully harness their hidden dynamic information.
\section{Acknowledgments}

J.S. is supported by the National Science Foundation Graduate Research Fellowship under Grant No. DGE-2146755. This work was supported by the U.S. Department of Energy, under DOE Contract No. DE-AC02-76SF00515. Flatiron Institute is a division of the Simons Foundation. Molecular graphics and analyses performed with UCSF ChimeraX, developed by the Resource for Biocomputing, Visualization, and Informatics at the University of California, San Francisco, with support from National Institutes of Health R01-GM129325 and the Office of Cyber Infrastructure and Computational Biology, National Institute of Allergy and Infectious Diseases.

\clearpage
\bibliographystyle{unsrt} 
\bibliography{ref.bib}

@article{meng2023ucsf,
  title={UCSF ChimeraX: Tools for structure building and analysis},
  author={Meng, Elaine C and Goddard, Thomas D and Pettersen, Eric F and Couch, Greg S and Pearson, Zach J and Morris, John H and Ferrin, Thomas E},
  journal={Protein Science},
  volume={32},
  number={11},
  pages={e4792},
  year={2023},
  publisher={Wiley Online Library}
}

@article{punjani2017cryosparc,
  title={cryoSPARC: algorithms for rapid unsupervised cryo-EM structure determination},
  author={Punjani, Ali and Rubinstein, John L and Fleet, David J and Brubaker, Marcus A},
  journal={Nature methods},
  volume={14},
  number={3},
  pages={290--296},
  year={2017},
  publisher={Nature Publishing Group}
}

@article{sledzieski2026landscape,
  title={The landscape of machine learning approaches for modeling protein conformational ensembles},
  author={Sledzieski, Samuel and Hanson, Sonya M},
  journal={Current Opinion in Structural Biology},
  volume={98},
  pages={103253},
  year={2026},
  publisher={Elsevier}
}

@article{harris2022selective,
  title={Selective G protein signaling driven by substance P--neurokinin receptor dynamics},
  author={Harris, Julian A and Faust, Bryan and Gondin, Arisbel B and D{\"a}mgen, Marc Andr{\'e} and Suomivuori, Carl-Mikael and Veldhuis, Nicholas A and Cheng, Yifan and Dror, Ron O and Thal, David M and Manglik, Aashish},
  journal={Nature chemical biology},
  volume={18},
  number={1},
  pages={109--115},
  year={2022},
  publisher={Nature Publishing Group US New York}
}

@article{campbell2020cryo,
  title={Cryo-EM reveals integrin-mediated TGF-$\beta$ activation without release from latent TGF-$\beta$},
  author={Campbell, Melody G and Cormier, Anthony and Ito, Saburo and Seed, Robert I and Bondesson, Andrew J and Lou, Jianlong and Marks, James D and Baron, Jody L and Cheng, Yifan and Nishimura, Stephen L},
  journal={Cell},
  volume={180},
  number={3},
  pages={490--501},
  year={2020},
  publisher={Elsevier}
}

@article{shimada2020structure,
  title={The structure of lipid nanodisc-reconstituted TRPV3 reveals the gating mechanism},
  author={Shimada, Hiroto and Kusakizako, Tsukasa and Dung Nguyen, TH and Nishizawa, Tomohiro and Hino, Tomoya and Tominaga, Makoto and Nureki, Osamu},
  journal={Nature Structural \& Molecular Biology},
  volume={27},
  number={7},
  pages={645--652},
  year={2020},
  publisher={Nature Publishing Group US New York}
}

@article{dang2019structural,
  title={Structural insight into TRPV5 channel function and modulation},
  author={Dang, Shangyu and van Goor, Mark K and Asarnow, Daniel and Wang, YongQiang and Julius, David and Cheng, Yifan and van der Wijst, Jenny},
  journal={Proceedings of the National Academy of Sciences},
  volume={116},
  number={18},
  pages={8869--8878},
  year={2019},
  publisher={National Academy of Sciences}
}

@article{culbertson2025cryo,
  title={Cryo-EM of human P-glycoprotein reveals an intermediate occluded conformation during active drug transport},
  author={Culbertson, Alan T and Liao, Maofu},
  journal={Nature Communications},
  volume={16},
  number={1},
  pages={3619},
  year={2025},
  publisher={Nature Publishing Group UK London}
}

@article{liebschner2019macromolecular,
  title={Macromolecular structure determination using X-rays, neutrons and electrons: recent developments in Phenix},
  author={Liebschner, Dorothee and Afonine, Pavel V and Baker, Matthew L and Bunk{\'o}czi, G{\'a}bor and Chen, Vincent B and Croll, Tristan I and Hintze, Bradley and Hung, L-W and Jain, Swati and McCoy, Airlie J and others},
  journal={Biological Crystallography},
  volume={75},
  number={10},
  pages={861--877},
  year={2019},
  publisher={International Union of Crystallography}
}

@article{Emsley:ba5144,
author = "Emsley, P. and Lohkamp, B. and Scott, W. G. and Cowtan, K.",
title = "{Features and development of {\it Coot}}",
journal = "Acta Crystallographica Section D",
year = "2010",
volume = "66",
number = "4",
pages = "486--501",
month = "Apr",
doi = {10.1107/S0907444910007493},
url = {https://doi.org/10.1107/S0907444910007493},
}

@article{Croll:ic5101,
author = "Croll, Tristan Ian",
title = "{{\it ISOLDE}: a physically realistic environment for model building into low-resolution electron-density maps}",
journal = "Acta Crystallographica Section D",
year = "2018",
volume = "74",
number = "6",
pages = "519--530",
month = "Jun",
doi = {10.1107/S2059798318002425},
url = {https://doi.org/10.1107/S2059798318002425},
}

@article{richman2025unlocking,
  title={Unlocking hidden biomolecular conformational landscapes in diffusion models at inference time},
  author={Richman, Daniel D and Karaguesian, Jessica and Suomivuori, Carl-Mikael and Dror, Ron O},
  journal={39th Conference on Neural Information Processing Systems (NeurIPS)},
  year={2025}
}

@article{chen2025building,
  title={Building molecular model series from heterogeneous CryoEM structures using Gaussian mixture models and deep neural networks},
  author={Chen, Muyuan},
  journal={Communications Biology},
  volume={8},
  number={1},
  pages={798},
  year={2025},
  publisher={Nature Publishing Group UK London}
}

@article{wohlwend2025boltz,
  title={Boltz-1 democratizing biomolecular interaction modeling},
  author={Wohlwend, Jeremy and Corso, Gabriele and Passaro, Saro and Getz, Noah and Reveiz, Mateo and Leidal, Ken and Swiderski, Wojtek and Atkinson, Liam and Portnoi, Tally and Chinn, Itamar and others},
  journal={BioRxiv},
  pages={2024--11},
  year={2025}
}

@article{tang2007eman2,
  title={EMAN2: an extensible image processing suite for electron microscopy},
  author={Tang, Guang and Peng, Liwei and Baldwin, Philip R and Mann, Deepinder S and Jiang, Wen and Rees, Ian and Ludtke, Steven J},
  journal={Journal of structural biology},
  volume={157},
  number={1},
  pages={38--46},
  year={2007},
  publisher={Elsevier}
}

@article{bakan2011prody,
  title={ProDy: protein dynamics inferred from theory and experiments},
  author={Bakan, Ahmet and Meireles, Lidio M and Bahar, Ivet},
  journal={Bioinformatics},
  volume={27},
  number={11},
  pages={1575--1577},
  year={2011},
  publisher={Oxford University Press}
}

@article{passaro2025boltz,
  title={Boltz-2: Towards accurate and efficient binding affinity prediction},
  author={Passaro, Saro and Corso, Gabriele and Wohlwend, Jeremy and Reveiz, Mateo and Thaler, Stephan and Somnath, Vignesh Ram and Getz, Noah and Portnoi, Tally and Roy, Julien and Stark, Hannes and others},
  journal={BioRxiv},
  year={2025}
}

@article{suzuki2026steering,
  title={Steering Conformational Sampling in Boltz-2 via Pair Representation Scaling},
  author={Suzuki, Shosuke and Amagasa, Toshiyuki},
  journal={bioRxiv},
  pages={2026--01},
  year={2026},
  publisher={Cold Spring Harbor Laboratory}
}

@article{Fadini2026,
  title = {AlphaFold as a prior: experimental structure determination conditioned on a pretrained neural network},
  ISSN = {1548-7105},
  url = {http://dx.doi.org/10.1038/s41592-026-03047-4},
  DOI = {10.1038/s41592-026-03047-4},
  journal = {Nature Methods},
  publisher = {Springer Science and Business Media LLC},
  author = {Fadini,  Alisia and Li,  Minhuan and McCoy,  Airlie J. and Banjara,  Suresh and Okumura,  Hiroki and Napier,  Eve and Fontana,  Pietro and Khan,  Amir R. and Jovine,  Luca and Terwilliger,  Thomas C. and Read,  Randy J. and Hekstra,  Doeke R. and AlQuraishi,  Mohammed},
  year = {2026},
  month = apr 
}

@article{raghu2025multiscale,
title       = {Multiscale guidance of protein structure prediction with heterogeneous cryo-EM data},
author      = {Rishwanth Raghu and Axel Levy and Gordon Wetzstein and Ellen D. Zhong},
journal   = {39th Conference on Neural Information Processing Systems (NeurIPS)},
year        = {2025}
}

@article{jamali2024automated,
  title={Automated model building and protein identification in cryo-EM maps},
  author={Jamali, Kiarash and K{\"a}ll, Lukas and Zhang, Rui and Brown, Alan and Kimanius, Dari and Scheres, Sjors HW},
  journal={Nature},
  volume={628},
  number={8007},
  pages={450--457},
  year={2024},
  publisher={Nature Publishing Group UK London}
}

@article{punjani20213d,
  title={3D variability analysis: Resolving continuous flexibility and discrete heterogeneity from single particle cryo-EM},
  author={Punjani, Ali and Fleet, David J},
  journal={Journal of structural biology},
  volume={213},
  number={2},
  pages={107702},
  year={2021},
  publisher={Elsevier}
}

@article{berman2000protein,
  title={The protein data bank},
  author={Berman, Helen M and Westbrook, John and Feng, Zukang and Gilliland, Gary and Bhat, Talapady N and Weissig, Helge and Shindyalov, Ilya N and Bourne, Philip E},
  journal={Nucleic acids research},
  volume={28},
  number={1},
  pages={235--242},
  year={2000},
  publisher={Oxford University Press}
}

@article{jumper2021highly,
  title={Highly accurate protein structure prediction with AlphaFold},
  author={Jumper, John and Evans, Richard and Pritzel, Alexander and Green, Tim and Figurnov, Michael and Ronneberger, Olaf and Tunyasuvunakool, Kathryn and Bates, Russ and {\v{Z}}{\'\i}dek, Augustin and Potapenko, Anna and others},
  journal={nature},
  volume={596},
  number={7873},
  pages={583--589},
  year={2021},
  publisher={Nature Publishing Group UK London}
}

@article{abramson2024accurate,
  title={Accurate structure prediction of biomolecular interactions with AlphaFold 3},
  author={Abramson, Josh and Adler, Jonas and Dunger, Jack and Evans, Richard and Green, Tim and Pritzel, Alexander and Ronneberger, Olaf and Willmore, Lindsay and Ballard, Andrew J and Bambrick, Joshua and others},
  journal={Nature},
  volume={630},
  number={8016},
  pages={493--500},
  year={2024},
  publisher={Nature Publishing Group UK London}
}

@article{zhong2021cryodrgn,
  title={CryoDRGN: reconstruction of heterogeneous cryo-EM structures using neural networks},
  author={Zhong, Ellen D and Bepler, Tristan and Berger, Bonnie and Davis, Joseph H},
  journal={Nature methods},
  volume={18},
  number={2},
  pages={176--185},
  year={2021},
  publisher={Nature Publishing Group US New York}
}

@article{cryodrgnai,
  title = "CryoDRGN-AI: neural ab initio reconstruction of challenging cryo-EM and cryo-ET datasets",
  author = "Levy, Axel and Raghu, Rishwanth and Feathers, Ryan and Grzadkowski, Michal and Poitevin, Frederic and
            Johnston, Jake D and Vallese, Francesca and Clarke, Oliver B and Wetzstein, Gordon and Zhong, Ellen D",
  journal = "Nature Methods",
  doi={10.1038/s41592-025-02720-4},
  url={nature.com/articles/s41592-025-02720-4},
  month={Jun},
  year={2025},
}

@article{lewis2025scalable,
  title={Scalable emulation of protein equilibrium ensembles with generative deep learning},
  author={Lewis, Sarah and Hempel, Tim and Jim{\'e}nez-Luna, Jos{\'e} and Gastegger, Michael and Xie, Yu and Foong, Andrew YK and Satorras, Victor Garc{\'\i}a and Abdin, Osama and Veeling, Bastiaan S and Zaporozhets, Iryna and others},
  journal={Science},
  pages={eadv9817},
  year={2025},
  publisher={American Association for the Advancement of Science}
}

@inproceedings{jing2024alphafold,
  title={AlphaFold Meets Flow Matching for Generating Protein Ensembles},
  author={Jing, Bowen and Berger, Bonnie and Jaakkola, Tommi},
  year={2024},
  booktitle={Forty-first International Conference on Machine Learning}
}

@article{wayment2024predicting,
  title={Predicting multiple conformations via sequence clustering and AlphaFold2},
  author={Wayment-Steele, Hannah K and Ojoawo, Adedolapo and Otten, Renee and Apitz, Julia M and Pitsawong, Warintra and H{\"o}mberger, Marc and Ovchinnikov, Sergey and Colwell, Lucy and Kern, Dorothee},
  journal={Nature},
  volume={625},
  number={7996},
  pages={832--839},
  year={2024},
  publisher={Nature Publishing Group UK London}
}

@article{monteiro2024high,
  title={High-throughput prediction of protein conformational distributions with subsampled AlphaFold2},
  author={Monteiro da Silva, Gabriel and Cui, Jennifer Y and Dalgarno, David C and Lisi, George P and Rubenstein, Brenda M},
  journal={nature communications},
  volume={15},
  number={1},
  pages={2464},
  year={2024},
  publisher={Nature Publishing Group UK London}
}

@article{lane2023protein,
  title={Protein structure prediction has reached the single-structure frontier},
  author={Lane, Thomas J},
  journal={Nature Methods},
  volume={20},
  number={2},
  pages={170--173},
  year={2023},
  publisher={Nature Publishing Group US New York}
}

@article{sala2023modeling,
  title={Modeling conformational states of proteins with AlphaFold},
  author={Sala, Davide and Engelberger, Felix and Mchaourab, Hassane S and Meiler, Jens},
  journal={Current Opinion in Structural Biology},
  volume={81},
  pages={102645},
  year={2023},
  publisher={Elsevier}
}

@article{ourmazd2022structural,
  title={Structural biology is solved—now what?},
  author={Ourmazd, Abbas and Moffat, Keith and Lattman, Eaton Edward},
  journal={Nature methods},
  volume={19},
  number={1},
  pages={24--26},
  year={2022},
  publisher={Nature Publishing Group US New York}
}

\appendix

\section{Technical appendices and supplementary material}

\subsection{Model Building Evaluation With Full MolProbity Metrics, Runtime, and Ablation Study}

We provide a complete version of the model building method comparison in Table \ref{table:model_building_metrics_full}, including additional information about all model geometry metrics (Ramachadran outliers, rotamer outliers, and clash scores), method runtimes, and an ablation of our structural loss. On top of the results shown in the main paper, we find that our method preserves valid stereochemistry Ramachandran and rotamer outliers compared to the unguided output of Boltz-2, and outputs structures with slightly higher clashes in order to achieve higher map--model fit. Runtimes are reported in terms of the average time taken to complete model building per map in the map series: since our method is trained jointly against all four maps per protein system (apart from the P-glycoprotein), we report the single map-averaged runtime in the table, finding that our method takes slightly longer than CryoBoltz in order to achieve higher accuracy.

The ablation results in the rightmost column highlight the importance of the structural loss $ \mathcal{L}_{\text{structural}} $: although removing this loss significantly improves map--model fit, it does so at the cost of poor stereochemical metrics, producing geometrically invalid structures. This experiment shows the importance of our structural loss at regularizing model building to output chemically valid conformations.

\begin{table}[t!]
  \caption{Model building performance across systems, with full model geometry metrics, method runtimes, and an ablation study on our method's structural loss ($ \mathcal{L}_{\text{structural}} $. Our method achieves competitive Ramachandran outlier and rotamer outlier scores, with slightly higher clash scores than other methods in some cases. While our method takes slightly longer than CryoBoltz and ModelAngelo, it achieves higher map--model fit as a result. The ablation study, reported in the rightmost column, demonstrates the importance of the structural loss at producing regularized atomic structures. *Method only produced structures for three out of the four maps.}
  \label{table:model_building_metrics_full}
  \centering
  \small
  \resizebox{\textwidth}{!}{
  \begin{tabular}{llccccc|c}
    \toprule
    & Metric & Boltz-2 \cite{passaro2025boltz} & CryoBoltz \cite{raghu2025multiscale} & E2GMM \cite{chen2025building} & ModelAngelo \cite{jamali2024automated} & Ours & Ours w/o Structural Loss \\
    \midrule
    \multirow{10}{*}{\rotatebox[origin=c]{90}{TRPV3}} & \textit{Map--Model Fit} &  &  &  &  &  &  \\
    & $CC_{volume}$ ($\uparrow$) & 0.31$\pm$0.03 & --- & 0.25$\pm$0.02 & 0.79$\pm$0.08 & 0.84$\pm$0.02 & 0.93$\pm$0.01 \\
    & $CC_{mask}$ ($\uparrow$) & 0.30$\pm$0.03 & --- & 0.23$\pm$0.02 & 0.79$\pm$0.07 & 0.84$\pm$0.02 & 0.93$\pm$0.01 \\
    & $CC_{peaks}$ ($\uparrow$) & 0.29$\pm$0.03 & --- & 0.21$\pm$0.02 & 0.71$\pm$0.08 & 0.79$\pm$0.02 & 0.95$\pm$0.01 \\
    & $CC_{box}$ ($\uparrow$) & 0.62$\pm$0.02 & --- & 0.55$\pm$0.01 & 0.85$\pm$0.04 & 0.89$\pm$0.01 & 0.95$\pm$0.01 \\
    \cmidrule{2-8}
    & \textit{Model Geometry} &  &  &  &  &  &  \\
    & Rama. Outlier (\%, $\downarrow$) & 0.00$\pm$0.00 & --- & 1.37$\pm$0.23 & 1.92$\pm$0.65 & 0.00$\pm$0.00 & 55.94$\pm$1.94 \\
    & Rota. Outlier (\%, $\downarrow$) & 0.06$\pm$0.08 & --- & 1.17$\pm$0.19 & 8.30$\pm$1.15 & 0.16$\pm$0.08 & 64.64$\pm$0.95 \\
    & Clash Score ($\downarrow$) & 21$\pm$2 & --- & 3$\pm$0 & 43$\pm$6 & 12$\pm$1 & 1661$\pm$17 \\
    \cmidrule{2-8}
    & Runtime (h:mm) & --- & --- & 5:39$\pm$0:12 & 0:07$\pm$0:01 & 0:33$\pm$0:00 & --- \\
    \midrule
    \multirow{10}{*}{\rotatebox[origin=c]{90}{P-glycoprotein}} & \textit{Map--Model Fit} &  &  &  &  &  &  \\
    & $CC_{volume}$ ($\uparrow$) & 0.39$\pm$0.04 & 0.73$\pm$0.03 & 0.26$\pm$0.02 & 0.59$\pm$0.15$^*$ & 0.77$\pm$0.02 & 0.96$\pm$0.01$^*$ \\
    & $CC_{mask}$ ($\uparrow$) & 0.40$\pm$0.03 & 0.74$\pm$0.03 & 0.26$\pm$0.03 & 0.59$\pm$0.14 & 0.77$\pm$0.02 & 0.96$\pm$0.01 \\
    & $CC_{peaks}$ ($\uparrow$) & 0.53$\pm$0.12 & 0.70$\pm$0.03 & 0.44$\pm$0.13 & -0.11$\pm$0.16 & 0.75$\pm$0.05 & 0.97$\pm$0.01 \\
    & $CC_{box}$ ($\uparrow$) & 0.62$\pm$0.07 & 0.85$\pm$0.02 & 0.54$\pm$0.08 & 0.33$\pm$0.09 & 0.85$\pm$0.02 & 0.96$\pm$0.01 \\
    \cmidrule{2-8}
    & \textit{Model Geometry} &  &  &  &  &  &  \\
    & Rama. Outlier (\%, $\downarrow$) & 0.05$\pm$0.04 & 3.72$\pm$6.02 & 1.29$\pm$0.34 & 2.13$\pm$3.80 & 0.11$\pm$0.07 & 62.50$\pm$14.87 \\
    & Rota. Outlier (\%, $\downarrow$) & 0.01$\pm$0.02 & 0.19$\pm$0.17 & 0.62$\pm$0.17 & 14.27$\pm$8.21 & 1.16$\pm$0.72 & 70.09$\pm$1.11 \\
    & Clash Score ($\downarrow$) & 21$\pm$4 & 152$\pm$37 & 4$\pm$1 & 70$\pm$43 & 109$\pm$44 & 2224$\pm$175 \\
    \cmidrule{2-8}
    & Runtime (h:mm) & --- & 0:15$\pm$0:01 & 3:39$\pm$0:05 & 0:06$\pm$0:00 & 0:46$\pm$0:00 & --- \\
    \midrule
    \multirow{10}{*}{\rotatebox[origin=c]{90}{Integrin $\alpha\text{V}\beta 8$}} & \textit{Map--Model Fit} &  &  &  &  &  &  \\
    & $CC_{volume}$ ($\uparrow$) & 0.62$\pm$0.07 & 0.38$\pm$0.13 & 0.35$\pm$0.05 & 0.58$\pm$0.03$^*$ & 0.68$\pm$0.05 & 0.74$\pm$0.04 \\
    & $CC_{mask}$ ($\uparrow$) & 0.61$\pm$0.06 & 0.39$\pm$0.12 & 0.29$\pm$0.06 & 0.57$\pm$0.02 & 0.68$\pm$0.05 & 0.75$\pm$0.04 \\
    & $CC_{peaks}$ ($\uparrow$) & 0.75$\pm$0.04 & 0.72$\pm$0.05 & 0.59$\pm$0.04 & -0.22$\pm$0.03 & 0.81$\pm$0.03 & 0.84$\pm$0.03 \\
    & $CC_{box}$ ($\uparrow$) & 0.75$\pm$0.04 & 0.72$\pm$0.05 & 0.59$\pm$0.04 & 0.28$\pm$0.02 & 0.81$\pm$0.03 & 0.84$\pm$0.03 \\
    \cmidrule{2-8}
    & \textit{Model Geometry} &  &  &  &  &  &  \\
    & Rama. Outlier (\%, $\downarrow$) & 0.02$\pm$0.03 & 1.64$\pm$0.46 & 2.26$\pm$0.43 & 16.59$\pm$2.73 & 0.02$\pm$0.03 & 1.01$\pm$0.76 \\
    & Rota. Outlier (\%, $\downarrow$) & 0.12$\pm$0.04 & 1.08$\pm$0.29 & 1.17$\pm$0.28 & 13.07$\pm$0.52 & 0.12$\pm$0.04 & 9.34$\pm$0.85 \\
    & Clash Score ($\downarrow$) & 19$\pm$3 & 419$\pm$25 & 6$\pm$1 & 97$\pm$19 & 19$\pm$3 & 220$\pm$46 \\
    \cmidrule{2-8}
    & Runtime (h:mm) & --- & 0:21$\pm$0:00 & 4:16$\pm$0:05 & 0:07$\pm$0:00 & 0:19$\pm$0:00 & --- \\
    \midrule
    \multirow{10}{*}{\rotatebox[origin=c]{90}{Neurokinin-1 GPCR}} & \textit{Map--Model Fit} &  &  &  &  &  &  \\
    & $CC_{volume}$ ($\uparrow$) & 0.55$\pm$0.06 & 0.73$\pm$0.05 & 0.75$\pm$0.04 & 0.69$\pm$0.06 & 0.73$\pm$0.04 & 0.80$\pm$0.03 \\
    & $CC_{mask}$ ($\uparrow$) & 0.55$\pm$0.06 & 0.72$\pm$0.05 & 0.75$\pm$0.04 & 0.70$\pm$0.06 & 0.72$\pm$0.04 & 0.79$\pm$0.02 \\
    & $CC_{peaks}$ ($\uparrow$) & 0.55$\pm$0.10 & 0.70$\pm$0.05 & 0.74$\pm$0.07 & 0.29$\pm$0.30 & 0.71$\pm$0.08 & 0.87$\pm$0.02 \\
    & $CC_{box}$ ($\uparrow$) & 0.75$\pm$0.04 & 0.84$\pm$0.03 & 0.85$\pm$0.02 & 0.59$\pm$0.22 & 0.83$\pm$0.03 & 0.86$\pm$0.02 \\
    \cmidrule{2-8}
    & \textit{Model Geometry} &  &  &  &  &  &  \\
    & Rama. Outlier (\%, $\downarrow$) & 0.00$\pm$0.00 & 0.84$\pm$0.22 & 2.31$\pm$0.74 & 5.15$\pm$2.03 & 0.00$\pm$0.00 & 24.97$\pm$2.14 \\
    & Rota. Outlier (\%, $\downarrow$) & 0.13$\pm$0.18 & 0.92$\pm$0.24 & 0.30$\pm$0.27 & 14.51$\pm$1.33 & 0.39$\pm$0.28 & 43.78$\pm$0.94 \\
    & Clash Score ($\downarrow$) & 22$\pm$2 & 157$\pm$13 & 6$\pm$2 & 80$\pm$17 & 23$\pm$3 & 872$\pm$29 \\
    \cmidrule{2-8}
    & Runtime (h:mm) & --- & 0:10$\pm$0:01 & 3:40$\pm$0:07 & 0:06$\pm$0:01 & 0:24$\pm$0:00 & --- \\
    \bottomrule
  \end{tabular}
  }
\end{table}

\subsection{Evaluating CryoSampler's Latent Space via Interpolation}

To evaluate the expressiveness of the latent space learned by our method's VAE, we conduct an additional experiment that trains on the two endpoint maps of each protein system's principal component studied in the main paper, rather than supervising on four maps. After training, we then sample eight intermediate atomic structures interpolated evenly between the latent codes for the input maps, and compute the ensemble accuracy of these interpolated structures against four intermediate maps along the same normal mode. As shown in Table \ref{table:ensemble_interpolation_metrics}, the interpolated structures slightly improve the ensemble fit over CryoSampler's two model-built structures (and greatly improve over Boltz-2 and normal mode analysis) in terms of precision, recall, and Wasserstein distance, suggesting that for individual protein systems, the latent space can properly interpolate atomic dynamics.

\begin{table}[h!]
  \caption{Evaluation of the latent space learned by our model. For each system, we select two maps corresponding to the endpoints of a normal mode extracted by 3DVA performed on the experimental data, and train CryoSampler on these maps for each system individually. We then sample 8 latent codes uniformly between the two codes corresponding to the endpoints (inclusive), and compute ensemble metrics via the $ CC_{volume} $ cross-correlation score in Phenix (described in Table \ref{table:ensemble_metrics}) against 4 intermediate experimental 3DVA maps along the same principal component. Our method is able to capture the intermediate map states to a higher degree of accuracy than Boltz-2 or normal mode analysis, demonstrating the interpolative capability of our generative model.}
  \label{table:ensemble_interpolation_metrics}
  \centering
  \small
  \begin{tabular}{llcccc}
    \toprule
    & Metric & Boltz-2 \cite{passaro2025boltz} & NMA \cite{bakan2011prody} & Ours (endpoints) & Ours (all states) \\
    \midrule
    \multirow{3}{*}{\rotatebox[origin=c]{90}{Integrin}} & Precision ($\uparrow$) & 0.68 & \underline{0.71} & \textbf{0.74} & \textbf{0.74} \\
    & Recall ($\uparrow$) & 0.68 & 0.69 & \underline{0.73} & \textbf{0.74} \\
    & $\mathcal{W}_2$ Dist. ($\downarrow$) & 0.33 & \underline{0.32} & \textbf{0.27} & \textbf{0.27} \\
    \midrule
    \multirow{3}{*}{\rotatebox[origin=c]{90}{NK-1}} & Precision ($\uparrow$) & 0.59 & 0.55 & \underline{0.69} & \textbf{0.70} \\
    & Recall ($\uparrow$) & 0.55 & 0.50 & \underline{0.62} & \textbf{0.65} \\
    & $\mathcal{W}_2$ Dist. ($\downarrow$) & 0.47 & 0.51 & \underline{0.38} & \textbf{0.37} \\
    \midrule
    \multirow{3}{*}{\rotatebox[origin=c]{90}{TRPV3}} & Precision ($\uparrow$) & 0.37 & 0.39 & \underline{0.84} & \textbf{0.85} \\
    & Recall ($\uparrow$) & 0.40 & 0.39 & \underline{0.84} & \textbf{0.85} \\
    & $\mathcal{W}_2$ Dist. ($\downarrow$) & 0.63 & 0.62 & \underline{0.16} & \textbf{0.15} \\
    \bottomrule
  \end{tabular}
\end{table}

\subsection{Cryo-EM Data Pre-processing}

In the following sub-sections, we describe the steps taken to process each cryo-EM dataset analyzed in the main text.

\subsubsection{TRPV3}

Picked particles from EMPIAR-10400 \cite{shimada2020structure} were downloaded and reconstructed via backprojection in CryoSPARC \cite{punjani2017cryosparc} with C4 symmetry. These were fed to a heterogeneous reconstruction job with 6 junk classes. The highest resolution class with 79,765 particles was subjected to homogeneous reconstruction with C4 symmetry, yielding a 3.2 angstrom reconstruction.

Heterogeneous reconstruction was subsequently performed with 3D Variability Analysis \cite{punjani20213d}, yielding five principal components with 20 maps each. We selected four maps evenly spaced along the fourth principal component, representing a ``breathing" motion of the protein. These maps were directly fed as input to each model building method without further pre-processing, using the \texttt{phenix.mtriage} command to additionally input estimated resolutions for each map to the individual algorithms. This command was used for all subsequent datasets except for the P-glycoprotein, for which per-map resolutions were available on the Electron Microscopy Data Bank.

\subsubsection{TRPV5}

Picked particles from EMPIAR-10255 \cite{dang2019structural} were cleaned for corrupt particles and subjected to ab initio reconstruction with a maximum resolution of 12 angstrom. Subsequent homogeneous refinement with C4 Symmetry resulted in a reconstruction of 3.0 angstrom. 3DVA \cite{punjani20213d} was used to perform heterogeneity analysis, producing five principal components with 20 maps each. Four evenly-spaced maps were selected along the fourth principal component, corresponding to a similar breathing motion as extracted for TRPV3. These maps were not used for model building, but rather as a held-out test set to evaluate the ensemble prediction capability (transferring conformations learned for TRPV3 to TRPV5).

\subsubsection{P-glycoprotein}

Four maps of the P-glycoprotein undergoing drug transport were downloaded from the Electron Microscopy Data Bank (\texttt{EMD-40226, 40259, 40258, 40227}) along with their corresponding deposited atomic models (\texttt{PDB: 8GMG, 8SA1, 8SA0, 8GMJ}) \cite{culbertson2025cryo}. These maps were then masked with a 5 angstrom extent around the PDB models using the \texttt{volume zone} command in ChimeraX \cite{meng2023ucsf} in order to remove background.

\subsubsection{Integrin $ \alpha V \beta 8 $}

The data were acquired from EMPIAR-10345 \cite{campbell2020cryo} and processed with CryoSPARC and 3DVA to obtain heterogeneous volumes. Four evenly-spaced maps were selected along the first principal component, and the volumes were resampled at 2.5 angstrom per voxel resolution and thresholded at minimum 0.28 intensity in ChimeraX. The density corresponding to the E and F chains of the protein were masked out using the \texttt{Segment Map} feature as Boltz-2 cannot accurately predict the binding of these dimers. Finally, the maps were masked to a tight box of size $ (50, 100, 90) $.

\subsubsection{Neurokinin-1 GPCR}

The data were acquired from EMPIAR-10786 \cite{harris2022selective} and processed with CryoSPARC and 3DVA to obtain heterogeneous volumes. Four evenly-spaced maps were selected along the first principal component, and the volumes were resampled at 2.5 angstrom per voxel resolution. The maps were masked with a 5 angstrom extent around the PDB model using the \texttt{volume zone} command, and floating density was removed using the command \texttt{surface dust size 5}.

\subsection{Settings for Baseline Methods and Evaluation}

All experiments were carried out on a single NVIDIA A100 GPU with 40 GB or 80 GB of memory (adjusted to the memory requirements of the evaluated method).

\subsubsection{Model Building}

Each baseline is run with default parameters on each of the maps separately over 3 replicates. Our method, E2GMM, and CryoBoltz require an initial structure, which we generate with Boltz-2 and align to the input map before performing model building. Sequences for each protein were acquired from the PDB \cite{berman2000protein}. In order to better match the structure output to the experimental density maps, the E and F chains for integrin were removed from the sequence input, and the A and B chains were truncated to only include the first 958 and 636 residues, respectively. Similarly, for the neurokinin-1 GPCR, the first 25 residues of the B chain were removed.

The atomic models output by each method were aligned to the corresponding map using the \texttt{fitmap} command in ChimeraX in order to remove any residual rigid body transformational error. B-factors were estimated for each model using the \texttt{phenix.real\_space\_refine} command in Phenix \cite{liebschner2019macromolecular} with the \texttt{run=adp} argument. The map--model fit and MolProbity geometry metrics were then computed using \texttt{phenix.map\_correlations} and \texttt{phenix.molprobity}.

\subsubsection{Ensemble Prediction}

We evaluated the ability of our method, Boltz-2, and normal mode analysis at generating ensembles of TRPV5. Our method was first pre-trained on the TRPV3 dataset. To train the diffusion MLP in the second stage, we augmented the four latent codes corresponding to the maps in the first stage with 93 latents interpolated evenly between each adjacent pair of true latents (exclusive). After diffusion training, our method was fed a static structure of TRPV5 from Boltz-2 and generated 16 conformations. Similarly, normal mode analysis in ProDy \cite{bakan2011prody} was used to generate 16 conformations along the first principal component, with modes computed using the $ C \alpha $ atoms with an RMSD extent of 1 angstrom. Boltz-2 was run 16 times independently to output 16 different conformations.

Unlike for model building, where we have access to corresponding pairs of atomic models and maps, the generated conformations in this stage are unpaired. As such, we computed pairwise analysis of every generated conformation from each method against each map in the held-out set for TRPV5. Each atomic conformation was aligned to each map using \texttt{fitmap} in ChimeraX, after which B-factors were computed using \texttt{phenix.real\_space\_refine}. Pairwise map--model fit was calculated using \texttt{phenix.map\_correlations}, and the resulting $ CC_{volume} $ metric was used to compute pairwise structure--map distances using the formula $ \sqrt{1 - CC_{volume}} $, from which precision, recall, and Wasserstein distances were calculated.


\end{document}